\documentclass[10pt,twocolumn,letterpaper]{article}

\usepackage{cvpr}              

\usepackage[dvipsnames]{xcolor}

\usepackage{multirow}
\usepackage{makecell}
\usepackage{bbding}
\usepackage{bm}
\usepackage{color}
\usepackage{colortbl}
\usepackage{adjustbox}

\definecolor{colorTab}{rgb}{0.95,0.91,0.93}

\definecolor{cvprblue}{rgb}{0.21,0.49,0.74}
\usepackage[pagebackref,breaklinks,colorlinks,citecolor=cvprblue]{hyperref}

\def\paperID{3579} 
\def\confName{CVPR}
\def\confYear{2024}

\title{Binarized Low-light Raw Video Enhancement}

\author {
    Gengchen Zhang $^{1}$, 
    Yulun Zhang $^{2}$,
    Xin Yuan $^{3}$,
    Ying Fu $^{1}$\thanks{Corresponding Author}\\
    \textsuperscript{\rm 1}Beijing Institute of Technology, \textsuperscript{\rm 2}Shanghai Jiao Tong University, \textsuperscript{\rm 3} Westlake University\\
    {\tt\small \{zhanggengchen,fuying\}@bit.edu.cn, yulun100@gmail.com, xylab@westlake.edu.cn}
}

\begin{document}
\maketitle
\begin{abstract}
Recently, deep neural networks have achieved excellent performance on low-light raw video enhancement. However, they often come with high computational complexity and large memory costs, which hinder their applications on resource-limited devices.
In this paper, we explore the feasibility of applying the extremely compact binary neural network (BNN) to low-light raw video enhancement. Nevertheless, there are two main issues with binarizing video enhancement models. One is how to fuse the temporal information to improve low-light denoising without complex modules. The other is how to narrow the performance gap between binary convolutions with the full precision ones. 
To address the first issue, we introduce a spatial-temporal shift operation, which is easy-to-binarize and effective. The temporal shift efficiently aggregates the features of neighbor frames and the spatial shift handles the misalignment caused by the large motion in videos.
For the second issue, we present a distribution-aware binary convolution, which captures the distribution characteristics of real-valued input and incorporates them into plain binary convolutions to alleviate the degradation in performance.
Extensive quantitative and qualitative experiments have shown our high-efficiency binarized low-light raw video enhancement method can attain a promising performance.
The code is available at \textcolor{magenta}{https://github.com/ying-fu/BRVE}.
\end{abstract}
\begin{figure}
  \begin{center}
  \includegraphics[width=\linewidth]{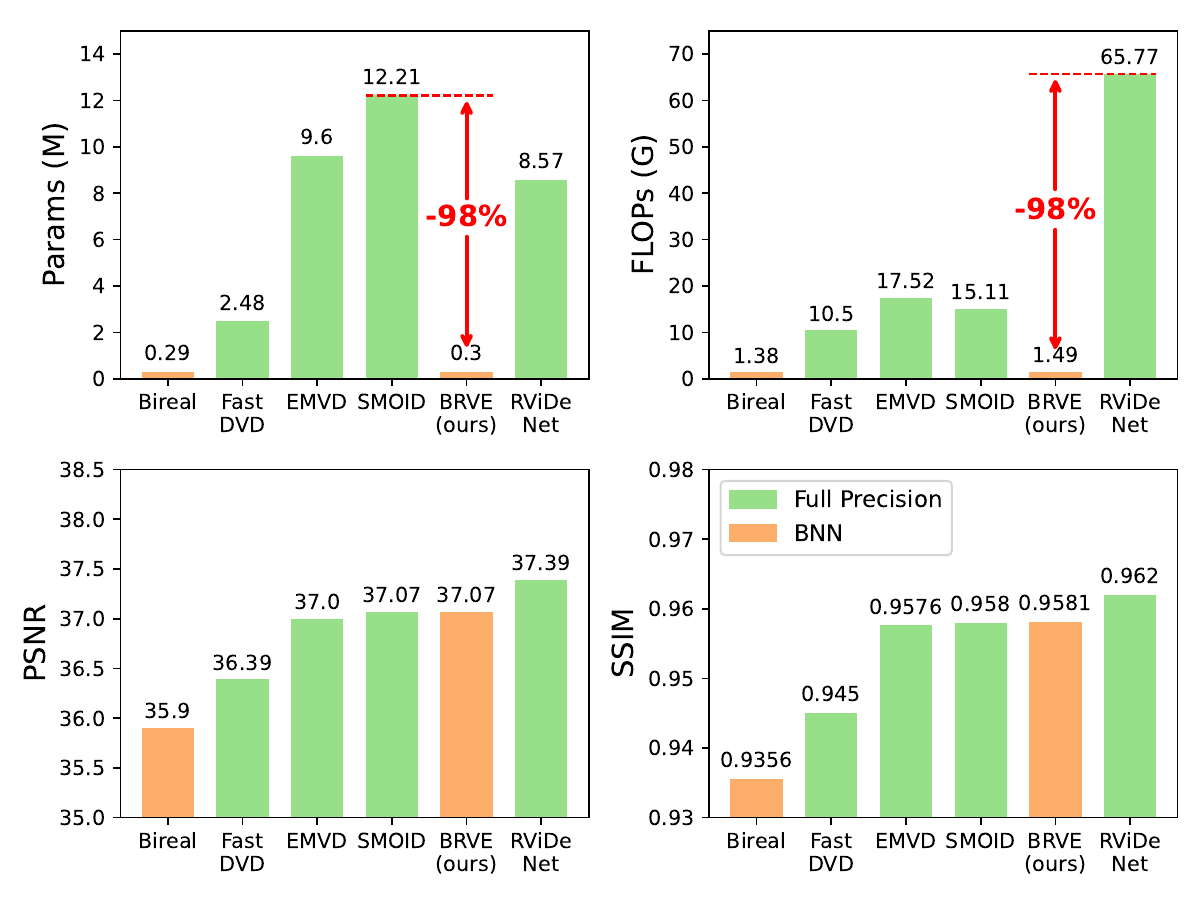}
  \end{center}
  \vspace{-7mm}
\caption{Efficiency and performance comparison of full precision networks and binary neural networks (BNNs).}
\label{fig:FP_vs_BNN}
  \vspace{-6mm}
\end{figure}

\vspace{-3mm}
\section{Introduction}
\vspace{-1mm}
Videos captured in low-light environments often suffer from degradations such as severe noise, color distortion, and lack of details. This not only causes poor video aesthetic quality but also impairs the application of video in downstream vision tasks \cite{LOD_BMVC_2021,LIS_IJCV_2023,NightDeraining_ICCV_2023}. To enhance low-light videos, there are several hardware-based solutions, including using high ISO, long exposure time, large aperture size, and a flashlight. 
However, these methods all have their own limitations.
For example, a high ISO setting amplifies noise, long exposure time causes motion blur in dynamic scenes, and flashlights have a limited range.
On the other hand, software-based low-light enhancement methods \cite{SID_CVPR_2018,SMID_ICCV_2019,SMOID_ICCV_2019,LEGAN_KBS_2022} can compensate for these limitations. Many works \cite{Huang_Towards_TIP_2022,LLRVD_TMM_2022,RAWHDR_ICCV_2023} have shown raw videos provide several advantages for low-light enhancement, such as linearity to scene lumination and high bit-depth to preserve more dark details.
In this work, we follow the raw-to-raw video enhancement pipeline to perform denoising on linearly scaled raw videos.

Deep learning methods have dominated low-light video enhancement \cite{SMID_ICCV_2019,SMOID_ICCV_2019} and denoising \cite{TaoZhang_ICCV_2021,TaoZhang_IJCV_2022,FloRNN_ECCV_2022,ShiftNet_CVPR_2023}. These deep models, such as convolutional neural networks (CNN) and transformers, often require high computational costs and large memory overhead. Nowadays, using mobile devices (\eg smartphones and hand-held cameras) to capture videos has become increasingly popular. However, these devices neither have the ability to directly acquire high-quality videos in low-light environments nor do they have enough resources to run deep neural networks.
To improve the efficiency of deep neural networks, many network compression techniques are proposed, including network quantization \cite{BNNSurvey_PR_2020,BNN_NIPS_2016}, parameter pruning \cite{pruning1,pruning2}, compact network design \cite{compact1}, and knowledge distillation \cite{distill1}. Among these approaches, binary neural network (BNN) stands out as an extreme case of network quantization, which binarizes both weights and activations (features) into only 1-bit (\ie -1 and +1) to reduce the memory overhead and uses bitwise operations to accelerate the computation.

Despite its superiority in efficiency, applying BNN to low-light video enhancement faces several challenges. 
\textit{\textbf{i)}} It is difficult for BNNs to ensure temporal consistency \cite{SMID_ICCV_2019,Zhang_CVPR_2021} to avoid flickering and effectively leverage spatial-temporal self-similarity, which is crucial for denoising \cite{VBM4D_2011,VNLB_2018}.
Some full precision methods use optical flow \cite{TOFlow_IJCV_2019,FloRNN_ECCV_2022} or deformable convolution \cite{RViDeNet_CVPR_2020} for feature alignment and fusion temporal information. However, they introduce auxiliary modules that are difficult to binarize.
Other methods directly employ 2D or 3D convolutions \cite{SMOID_ICCV_2019,FastDVD_CVPR_2020} for implicit alignment, which has a limited receptive field to deal with large motions in videos.
\textit{\textbf{ii)}} Binary convolutions encounter degradation in representation capability compared to their full precision counterparts \cite{Bireal_ECCV_2018}. Because binarizing real-valued activations and weights to 1-bit leads to information loss of absolute value and distribution characteristics.

In this paper, we design a binary raw video enhancement model (BRVE) to address these issues. 
Specifically, we introduce an efficient spatial-temporal shift operation to fully exploit the temporal redundancy for video enhancement. We use a cyclic temporal shift to fuse the features of frames in a local window and perform a spatial shift on features to enlarge the receptive field and handle the misalignment. These shift operations do not introduce extra parameters and enable parallel processing of multiple frames using 2D convolutions.
Besides, we propose a distribution-aware binary convolution that can reduce the performance gap between binary convolutions and full precision ones. It employs a distribution-aware channel attention to extract a real-valued scale factor from the input activation with negligible computation. The scale factor injects distribution information of full precision input into vanilla binary convolutions to reduce the quantization error. 
As shown in Figure \ref{fig:FP_vs_BNN}, our binarized low-light video enhancement method can reduce the model size and computation while having a comparable performance with full precision ones.
Our main contributions can be summarized as follows:
\vspace{-1mm}

\begin{itemize}[noitemsep,topsep=2pt]
    \item We build a compact binarized model for low-light raw video enhancement, which can achieve satisfactory results with low memory and computation.
    \vspace{1mm}
    \item We design an easy-to-binarize spatial-temporal shift operation to tackle the alignment of features and aggregate temporal information for low-light video enhancement.
    \vspace{1mm}
    \item We propose a distribution-aware binary convolution to mitigate the information loss of real-valued activations caused by binarization.
\end{itemize}
\vspace{-1mm}
\section{Related Work}
\vspace{-2mm}

In this section, we review datasets and methods for low-light video enhancement and various applications of BNNs.

\subsection{Low-light Video Enhancement}
Compared to low-light image enhancement, enhancing low-light videos is more challenging because obtaining paired low-light and clean, noise-free normal exposure videos is difficult. 
Therefore, some work uses camera noise model \cite{Wang_ICCV_2019, ELD_CVPR_2020, Zhang_CVPR_2021,Zou_CVPR_2022,ELD_PAMI_2022} or Generative Adversarial Networks (GAN) \cite{SIDGAN_ECCV_2020} to synthesize low-light noisy videos for training neural networks. For real-captured training data, Chen \etal \cite{SMID_ICCV_2019} collects a static Dark Raw Video (DRV) dataset.
To obtain paired low/normal-light videos with motion, some work adopts mechatronic systems \cite{SDSD_ICCV_2021,DID_ICCV_2023} to reproduce the same motion twice.
Jiang \etal \cite{SMOID_ICCV_2019} builds a dual-camera system to shoot video pairs at once. A beam splitter is used to generate two identical light beams and one of them is weakened through a neutral density (ND) filter. 
To get rid of extra equipment, Fu \etal \cite{LLRVD_TMM_2022} shoots a sequence of dark/bright frame pairs on a 4K monitor to form a video pair. 
Among these datasets, we mainly focus on enhancing low-light raw videos. Because raw data is directly obtained from the sensor and is not processed by non-linear operations in the image signal processor (ISP). We use the linearly scaling then denoising pipeline for low-light raw video enhancement following previous methods \cite{SID_CVPR_2018,SMID_ICCV_2019,SMOID_ICCV_2019}.

\begin{figure*}
  \begin{center}
  \includegraphics[width=\linewidth]{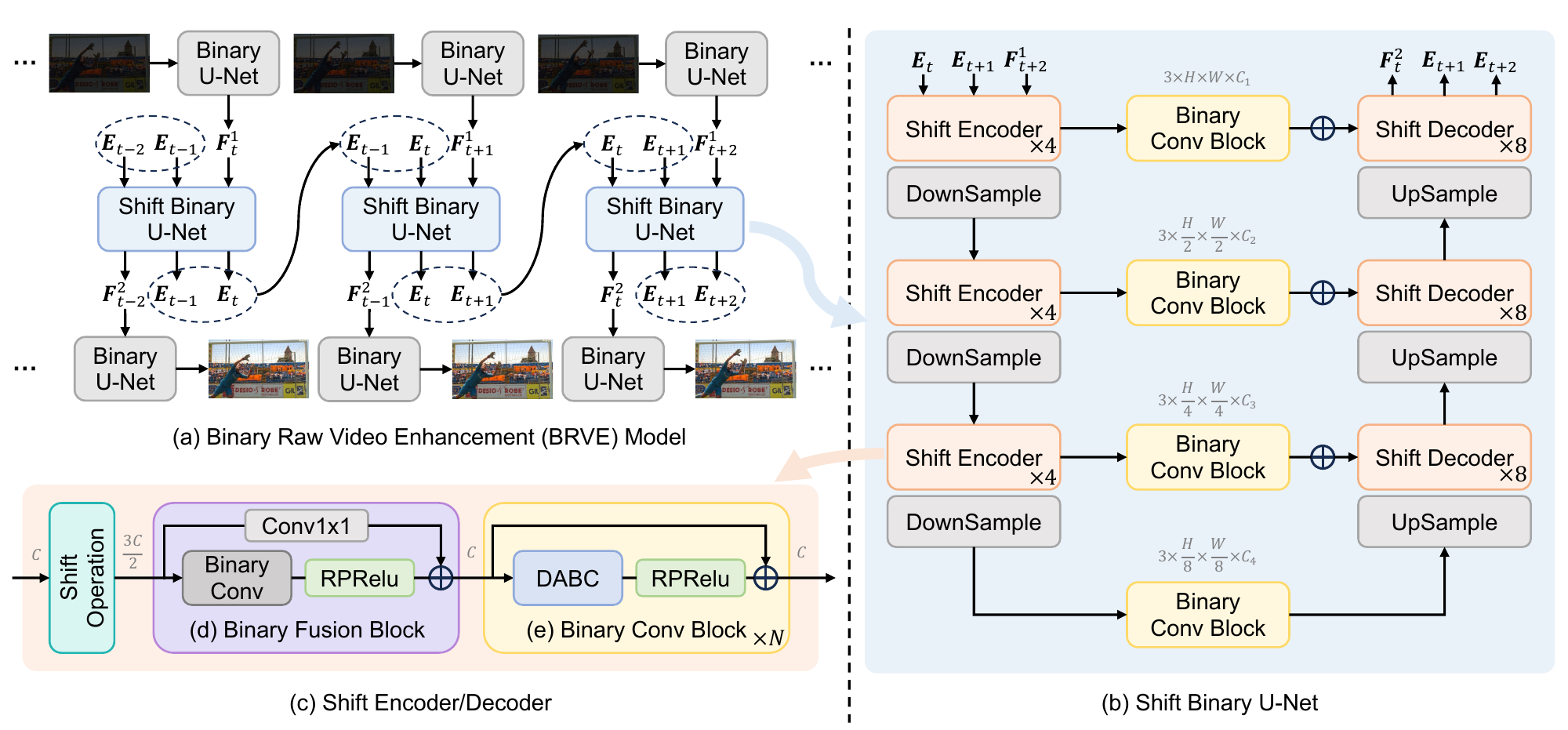}
  \end{center}
  \vspace{-5mm}
\caption{Overall architecture of BRVE model. (a) BRVE uses a shift binary U-Net for local feature fusion and exploits recurrent embeddings for long-range feature propagation. (b) Shift binary U-Net. (c) Shift encoder/decoder consists of a shift operation, a binary fusion block, and several binary conv blocks using the distribution-aware binary convolution (DABC).}
\label{fig:framework}
\end{figure*}

Directly using low-light image enhancement methods without considering temporal consistency may cause a flickering problem. Besides, processing low-light video frame-wisely leads to inferior denoising performance because the temporal redundancy is not fully exploited.
Recently, many deep-learning models have been developed to improve temporal consistency and make use of spatial-temporal information for low-light video enhancement and denoising \cite{SMOID_ICCV_2019,LLRVD_TMM_2022,RDRF_ACMMM_2023,3D2UNet_PRCV_2021}. 
SMOID \cite{SMOID_ICCV_2019} exploits 3D convolutions to aggregate temporal features for video enhancement. 
Some methods like DVDNet \cite{DVDNet_ICIP_2019} and TOFlow \cite{TOFlow_IJCV_2019} use optical flow to align neighbor frames or features for denoising. RViDeNet \cite{RViDeNet_CVPR_2020} performs alignment with pyramidal deformable convolution for raw video denoising. ShiftNet \cite{ShiftNet_CVPR_2023} adopts grouped spatial-temporal shift to simplify feature aggregation. In order to improve the efficiency of the networks, FastDVD \cite{FastDVD_CVPR_2020} uses two-stage cascaded U-Nets for implicit motion compensation. 
EMVD \cite{EMVD_CVPR_2020} leverages a fusion stage to efficiently reduce noise with former frames. 
However, these methods only focus on lightweight network design without exploring hardware-friendly approaches like model quantization and binarization.

\subsection{Binary Neural Networks}
Among the various deep neural network compression techniques, binary neural network (BNN) is an extreme form of network quantization. In a BNN, the network's weight parameters and activation values are represented using 1-bit values (\ie -1 and +1) which can significantly reduce the storage and computational requirements \cite{BNNSurvey_PR_2020}. The pioneering work \cite{BinaryConnect_NIPS_2015, BNN_NIPS_2016} uses the sign function to obtain binarized activations and weights and optimize network parameters with the straight-through-estimator (STE). However, BNNs face the loss of precision due to quantization error, gradient error, and limited representation ability. To reduce the quantization error, XNOR-Net \cite{XNORNet_ECCV_2016} utilizes scaling factors for weights and activations. IRNet \cite{IRNet_CVPR_2020} proposes to normalize the weights to retain information and increase information entropy. To reduce the gradient error, many efforts have been made to optimize the approximation of the sign function. Liu \etal \cite{Bireal_ECCV_2018} uses a piecewise quadratic function. Qin \etal \cite{IRNet_CVPR_2020} introduces an error decay estimator (EDE) with the tanh function. Xu \etal \cite{FDA_NIPS_2021} proposes to decompose the sign function with Fourier series. 
Wu \etal \cite{ReSTE_ICCV_2023} develops a Rectified STE (ReSTE) to balance the gradient estimating error and the training stability. 
To improve the capability of BNNs, Bi-real \cite{Bireal_ECCV_2018} preserves real-weight activations through a simple shortcut. ReActNet \cite{ReActNet_ECCV_2020} and INSTA-BNN \cite{INSTA_ICCV_2023} learn a channel-wise threshold to improve the binary activation function for activations.

In addition to its widespread use in high-level vision tasks, BNNs are also applied in low-level vision \cite{BNNSR_ECCV_2020,BTM_AAAI_2021,BBCU_ICLR_2023,BSCI_arxiv_2023}. Jiang \etal \cite{BTM_AAAI_2021} removes batch normalization (BN) layers and introduces a binary training mechanism for image super-resolution. Xia \etal \cite{BBCU_ICLR_2023} designs a basic binary convolution unit (BBCU) for image restoration. Cai \etal \cite{BSCI_arxiv_2023} proposes a BiSRNet for hyperspectral image reconstruction. Nevertheless, how to use binary networks to extract temporal information in videos has not been discussed.

\section{Method}

In this section, we first introduce our low-light raw video enhancement pipeline. Then we present the architecture of our binary raw video enhancement (BRVE) model.
Finally, we elaborate on the details of the distribution-aware binary convolution and spatial-temporal shift operation.

\subsection{Binary Raw Video Enhancement Model}

\paragraph{Problem Formulation.} Given a consecutive sequence of low-light noisy raw video frames in Bayer pattern, denoted as $\{\bm{I}^{B}_{1}, \bm{I}^{B}_{2}, ... , \bm{I}^{B}_{T} \}$, where $\bm{I}^{B}_{i} \in \mathbb{R}^{H \times W}$ and $T$ is the number of input frames. Following previous work \cite{SID_CVPR_2018,RViDeNet_CVPR_2020}, we first pack each $2 \times 2$ Bayer pattern into four color channels. Benefiting from the linear relationship of raw data with exposure levels, we utilize a predefined scaling factor $r$ to amplify low-light videos to a proper brightness. These packed and amplified frames are denoted as $\{\bm{I}^{P}_{1}, \bm{I}^{P}_{2}, ... , \bm{I}^{P}_{T} \}$, where $\bm{I}^{P}_{i} \in \mathbb{R}^{\frac{H}{2} \times \frac{W}{2} \times 4}$, which are then fed into the BRVE model to generate clean bright frames $\{\bm{O}^{P}_{1}, \bm{O}^{P}_{2}, ... , \bm{O}^{P}_{T} \}$. \vspace{-5mm}

\paragraph{Overall Architecture.} 
The overall architecture of our BRVE model is illustrated in Figure \ref{fig:framework} \textcolor{red}{(a)}.
In the first stage, we use a binary U-Net \cite{UNet_MICCAI_2015} to extract features for each input raw frame. 
The binary U-Net uses binary conv blocks as encoders and decoders. The structure of binary conv blocks is shown in Figure \ref{fig:framework} \textcolor{red}{(e)}. It adopts the distribution-aware binary convolution (DABC) to improve the capability of BNNs, which will be introduced in Section \ref{sec:DABC}.
As the high system gain in low-light photography and the amplification factor increase the noise, the first U-Net can also serve as a pre-denoising step. 
As shown in Figure \ref{fig:framework} \textcolor{red}{(b)}, the main component of BRVE is a shift U-Net, which expoit shift encoders/decoders in each level. As illustrated in Figure \ref{fig:framework} \textcolor{red}{(c)}, it applies shift operation for spatial-temporal feature fusion in a local sliding window, which will be discussed in Section \ref{sec:Shift}.
In the last stage, we also use a binary U-Net to aggregate features in the former two stages for raw video reconstruction. 
Following previous BNN work \cite{Bireal_ECCV_2018,BBCU_ICLR_2023}, we do not binarize the first convolution in the first U-Net to retain more information of raw input and the last convolution in the last U-Net to better maps features to bright and denoised raw frames. \vspace{-4mm}

\paragraph{Recurrent Embedding.} As shown in Figure \ref{fig:framework} \textcolor{red}{(a)}, we add recurrent embeddings between neighbor windows for long-range feature propagation. For a local temporal window $\{ t-1, t, t+1 \}$, the shift binary U-Net takes two embedding features $\bm{E}_{t-1}, \bm{E}_{t}$ from previous window and $\bm{F}^{1}_{t+1}$ as input. The first output $\bm{F}^{2}_{t-1}$ is then fed into the next stage and the latter two outputs $\bm{E}_{t}, \bm{E}_{t+1}$ are used as recurrent embeddings for next window.

\subsection{Distribution-Aware Binary Convolution}\label{sec:DABC}

The main reason for the performance degradation of binary convolutions is the loss of full precision information in the convolution kernel weights and activations. To address this issue, real-valued scale factors \cite{XNORNet_ECCV_2016,Xu_Accurate_BMVC_2019} are often adopted to multiply with the output of binary convolution. However, most of them are fixed during inference making it less flexible to use the information of different real-valued input activations.
To fully exploit full precision information, we propose the distribution-aware binary convolution (DABC) that uses distribution-aware channel attention (DACA) to efficiently generate dynamic scale factors in terms of the distribution characteristic of input activations. \vspace{-5mm}

\begin{figure}
  \begin{center}
  \includegraphics[width=\linewidth]{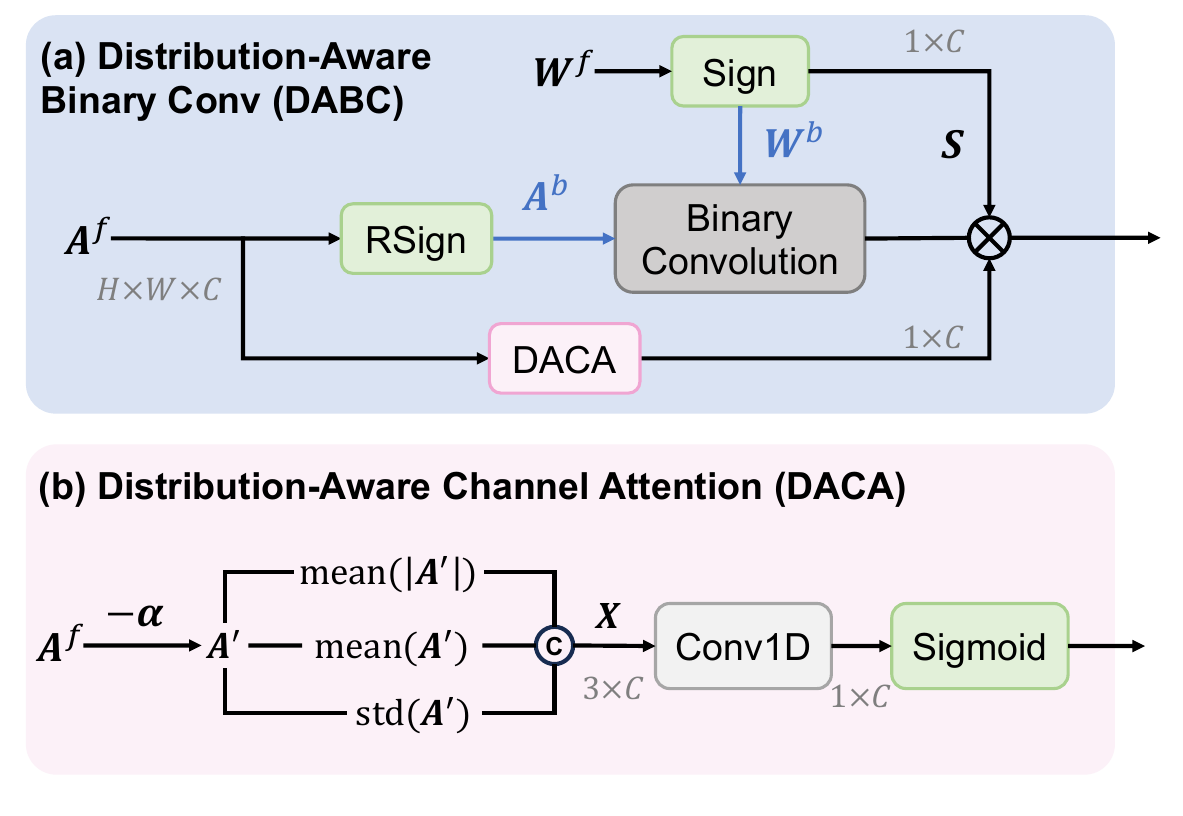}
  \end{center}
  \vspace{-6mm}
\caption{Distribution-Aware Binary Convolution (DABC).}
\label{fig:BinaryConv}
\vspace{-4mm}
\end{figure}

\paragraph{Binary Convolution.} The core of the binary convolution is to binarize full precision activation $\bm{A}^{f} \in \mathbb{R}^{H \times W \times C}$ and kernel weight $\bm{W}^{f} \in \mathbb{R}^{C_{out} \times C \times K \times K}$ to a binary set $\mathbb{B} = \{ -1, +1 \}$. As shown in Figure \ref{fig:BinaryConv} \textcolor{red}{(a)}, the binarize procedure of the weight can be represented as
\begin{equation}
\begin{split}
\bm{W}^{b} &= {\rm Sign}(\bm{W}^{f}) =
\left\{
   \begin{array}{lr}
   +1, \ \ \bm{W}^{f} > 0 \\
   -1, \ \ \bm{W}^{f} \leq 0
   \end{array}
\right. , \\
\bm{S}_{i} &= \frac{|| \bm{W}^{f}_{i} ||_{1}}{C \times K \times K}, \ i = 1 ... C_{out},
\end{split}
\label{eq:SignW}
\end{equation}
where $\bm{W}^{b} \in \mathbb{B}^{C_{out} \times C \times K \times K}$ is the binary kernel weight. 
Following previous work \cite{XNORNet_ECCV_2016}, the scale factor $\bm{S} \in \mathbb{R}^{C_{out}}$ is used for reducing the quantization error of the binary kernel weight. The real-valued activation is binarized by the RSign \cite{ReActNet_ECCV_2020} function, \ie,
\begin{equation}
\bm{A}^{b} = {\rm RSign}(\bm{A}^{f}) = {\rm Sign}(\bm{A}^{'}),
\label{eq:SignA}
\end{equation}
where $\bm{A}^{b} \in \mathbb{B}^{H \times W \times C}$ is the binarized activation and $\bm{A}^{'} = \bm{A}^{f} - \bm{\alpha}$, where $\bm{\alpha} \in \mathbb{R}^{C}$ is a learnable channel-wise parameter that controls the threshold for binarizing the input activation. With binary activation and weight, the binary convolution $\otimes$ can be efficiently computed using bitwise operations
\begin{equation}
\bm{A}^{b} \otimes \bm{W}^{b} = {\rm bitcount}({\rm XNOR}(\bm{A}^{b}, \bm{W}^{b})).
\label{eq:BinaryConv}
\end{equation} \vspace{-8mm}

\paragraph{Distribution-Aware Channel Attention.} To reduce the information loss from real-valued activation, we first collect several channel-wise statistics from $\bm{A}^{'}$, which describes the distribution characteristic of the activation \cite{INSTA_ICCV_2023}. Inspired by the efficient channel attention module \cite{ECA_CVPR_2020}, we use a 1D convolution followed by a Sigmoid function to extract a distribution-aware scale factor from these statistics with negligible computation and parameters. As shown in Figure \ref{fig:BinaryConv} \textcolor{red}{(b)}, the DACA module can be represented as
\begin{equation}
\begin{split}
\bm{X} = {\rm Concat}({\rm mean}(|\bm{A}^{'}|),& {\rm mean}(\bm{A}^{'}), {\rm std}(\bm{A}^{'})), \\
{\rm DACA}(\bm{A}^{'}) = {\rm Sigmoid}&({\rm Conv1d}(\bm{X})),
\end{split}
\label{eq:DACA}
\end{equation}
where $\bm{X} \in \mathbb{R}^{C \times 3}$ and ${\rm DACA}(\bm{A}^{'}) \in \mathbb{R}^{C \times 1}$. Finally, the distribution-aware binary convolution is defined as
\begin{equation}
\begin{split}
\bm{Y} = (\bm{A}^{b} \otimes \bm{W}^{b}) \odot \bm{S} \odot {\rm DACA}(\bm{A}^{'}) 
\end{split}
\label{eq:DABC}
\end{equation}
where $\bm{Y} \in \mathbb{Z}^{H \times W \times C_{out}}$ is the output activation rescaled with the factors from weights and DACA module. \vspace{-5mm}

\paragraph{Basic Binarized Modules.} As shown in Figure \ref{fig:framework} \textcolor{red}{(e)}, a binary convolution and a RPRelu \cite{ReActNet_ECCV_2020} function form the binary conv block. RPRelu function can be formulated as
\begin{equation}
\begin{split}
{\rm RPRelu}(\bm{Y}) = 
\left\{
   \begin{array}{lr}
   \bm{Y}_{i} - \bm{\gamma}_{i} + \bm{\zeta}_{i}, \ \ \bm{Y}_{i} > \bm{\zeta}_{i} \\
   \bm{\beta}_{i}(\bm{Y}_{i} - \bm{\gamma}_{i}) + \bm{\zeta}_{i}, \ \ \bm{Y}_{i} \leq \bm{\zeta}_{i},
   \end{array}
\right.
\end{split}
\label{eq:RPRlue}
\end{equation}
where $\bm{\gamma}_{i}, \bm{\zeta}_{i}, \bm{\beta}_{i} \in \mathbb{R}^{C_{out}}$ are learnable parameters. As incorporating a full precision shortcut connection is crucial in binary convolution to compensate for the information loss, we add the full precision input $\bm{A}^{f}$ with the output of binary convolution following previous work \cite{Bireal_ECCV_2018,BBCU_ICLR_2023}. However, binary convolution needs to change the channel number of activation (\ie $C \neq C_{out}$) in the shift encoder/decoder, which disables the addition of shortcut as well as the multiplication of DACA module. As shown in Figure \ref{fig:framework} \textcolor{red}{(d)}, we use a full precision $1 \times 1$ convolution to replace them to match the channel in the binary fusion block. To minimize computation cost, the binary fusion block is only used once in each shift encoder/decoder. We use bilinear upsampling and average pooling for downsampling in the binary U-Net.

\subsection{Spatial-Temporal Shift Operation}\label{sec:Shift}

\begin{figure}
  \begin{center}
  \includegraphics[width=\linewidth]{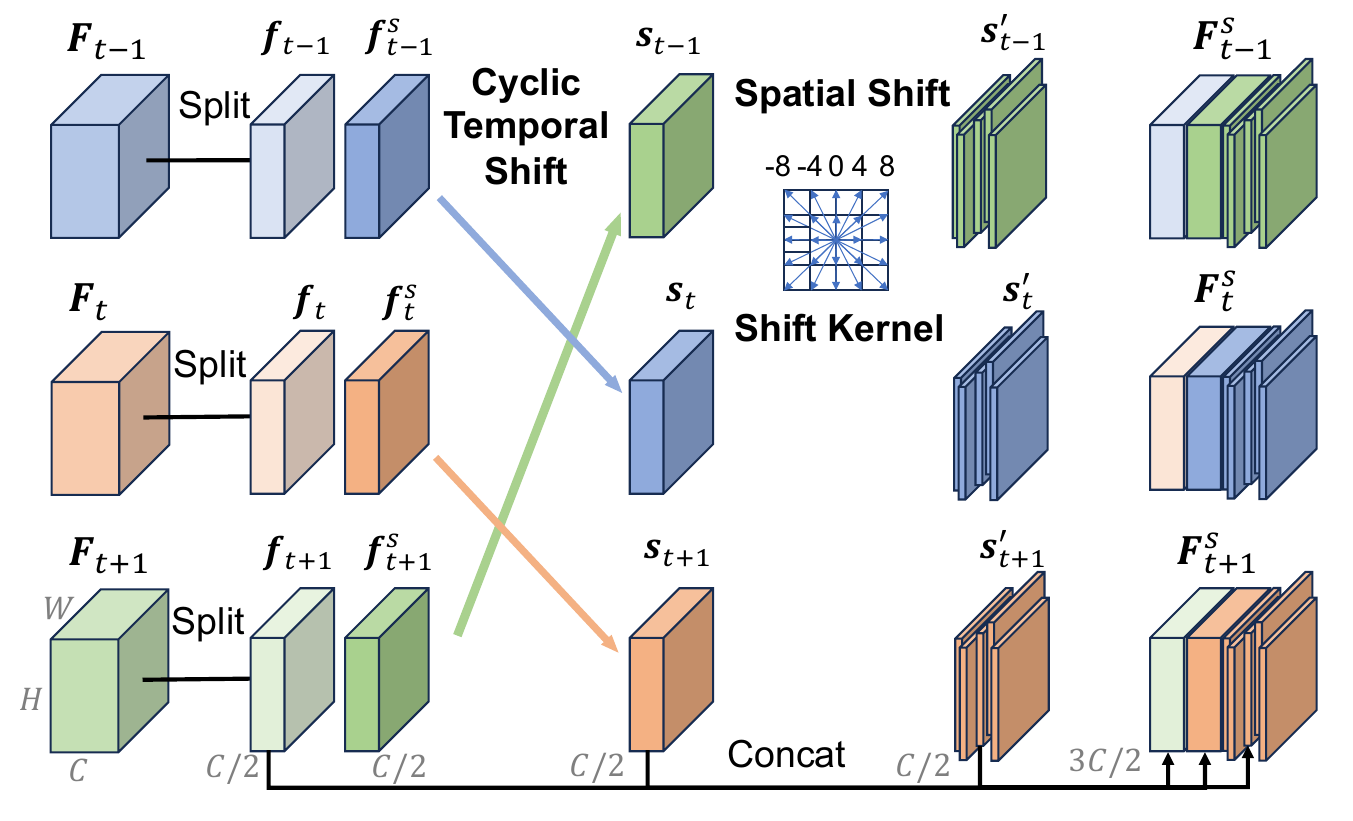}
  \end{center}
    \vspace{-5mm}
\caption{Spatial-temporal shift operation.}
\label{fig:Shift}
\vspace{-1mm}
\end{figure}

Spatial-temporal self-similarity is a crucial property for video denoising. Previous methods either use complicated models that are difficult to quantize \cite{TOFlow_IJCV_2019,FloRNN_ECCV_2022,RViDeNet_CVPR_2020} or directly use convolution \cite{FastDVD_CVPR_2020} for temporal feature fusion. Inspired by the ShiftNet \cite{ShiftNet_CVPR_2023}, we introduce a temporal-spatial shift operation, which propagates information between neighbor frames through the temporal shift and enlarges the receptive field of the network with the spatial shift. The shift operation is performed at the beginning of each shift encoder/decoder as shown in Figure \ref{fig:framework} \textcolor{red}{(c)}. It is efficient and parameter-free because it only transforms the input features and uses subsequent binarized modules described in Section \ref{sec:DABC} for feature fusion. \vspace{-5mm}

\paragraph{Cyclic Temporal Shift.} In order to enable bi-directional feature propagation, ShiftNet \cite{ShiftNet_CVPR_2023} performs forward and backward temporal shift alternatively in a sequence of features. However, caching these features requires a significant storage overhead and prevents online processing \cite{FloRNN_ECCV_2022}. We propose a cyclic temporal shift to aggregate features in a three-frame local window to reduce the feature cache.

As shown in Figure \ref{fig:Shift}, the inputs of shift operation are three consecutive features $\{ \bm{F}_{t-1}, \bm{F}_{t}, \bm{F}_{t+1} \}$. Each of them is first split into two parts along the channel dimension 
\begin{equation}
  [ \bm{f}_{i}, \bm{f}^{s}_{i} ] = {\rm Split}(\bm{F}_{i}), \  i \in \{ t-1, t, t+1 \},
\label{eq:split}
\end{equation}
where $\bm{F}_{i} \in \mathbb{R}^{H \times W \times C}$ and $\bm{f}_{i}, \bm{f}^{s}_{i} \in \mathbb{R}^{H \times W \times C/2}$. The first part $\bm{f}_{i}$ keeps information of frame $i$ and the second part $\bm{f}^{s}_{i}$ is passed to the neighbor frame to perform spatial-temporal shift and obtain $\bm{F}^{s}_{i}$ for feature fusion.

In cyclic temporal shift, the second part features $\bm{f}^{s}_{t-1}, \bm{f}^{s}_{t}$ shift to next position and feature $\bm{f}^{s}_{t+1}$ becomes the first one, as shown in Figure \ref{fig:Shift}. After the temporal shift, we can obtain a reordered set of features
\begin{equation}
  \begin{split}
  S^{t} = \{ \bm{s}_{t-1}, \bm{s}_{t}, \bm{s}_{t+1} \} = \{ \bm{f}^{s}_{t+1}, \bm{f}^{s}_{t-1}, \bm{f}^{s}_{t} \}.
  \end{split}
\label{eq:TempShift}
\end{equation}
However, performing the cyclic temporal shift operation once is insufficient to integrate each feature with the other two features in the local window. We also shift the features in the opposite direction along the temporal dimension and the shifted set $S^{t}$ becomes $\{ \bm{f}^{s}_{t}, \bm{f}^{s}_{t+1}, \bm{f}^{s}_{t-1} \}$. In each level of shift binary U-Net, we conduct cyclic temporal shift for two directions alternatively to fuse the features adequately. \vspace{-3mm}

\begin{table*}
\renewcommand\arraystretch{1}
\begin{center}
\setlength{\tabcolsep}{0.6mm} 
\begin{tabular}{l|ccc|ccc|ccc|c|c}
\hline
\multicolumn{1}{c|}{\multirow{2}{*}{Method}} & \multicolumn{3}{c|}{Gain0} & \multicolumn{3}{c|}{Gain15} & \multicolumn{3}{c|}{Gain30} & \multicolumn{1}{c|}{\multirow{2}{*}{\makecell[c]{Params \\ (M)}}} & \multicolumn{1}{c}{\multirow{2}{*}{\makecell[c]{FLOPs \\ (G)}}} \\
\cline{2-10}
\multicolumn{1}{c|}{} & \multicolumn{1}{c}{PSNR $\uparrow$} & \multicolumn{1}{c}{SSIM $\uparrow$} & \multicolumn{1}{c|}{ST-RRED $\downarrow$} & \multicolumn{1}{c}{PSNR $\uparrow$} & \multicolumn{1}{c}{SSIM $\uparrow$} & \multicolumn{1}{c|}{ST-RRED $\downarrow$} & \multicolumn{1}{c}{PSNR $\uparrow$} & \multicolumn{1}{c}{SSIM $\uparrow$} & \multicolumn{1}{c|}{ST-RRED $\downarrow$} & \multicolumn{1}{c|}{} & \multicolumn{1}{c}{} \\
\hline
SMOID \cite{SMOID_ICCV_2019} & 39.74 & 0.9733 & 0.0446 & 39.43 & 0.9751 & 0.0461 & 40.06 & 0.9769 & 0.0431 & 12.21 & 15.11\\
RViDeNet \cite{RViDeNet_CVPR_2020} & 41.09 & 0.9778 & 0.0510 & 40.94 & 0.9803 & 0.0494 & 41.60 & 0.9818 & 0.0500 & 8.57 & 65.77\\
FastDVD \cite{FastDVD_CVPR_2020} & 40.37 & 0.9724 & 0.0765 & 40.66 & 0.9768 & 0.0664 & 40.59 & 0.9763 & 0.1812 & 2.48 & 10.50\\
EMVD-L \cite{EMVD_CVPR_2020} & 41.10 & 0.9785 & 0.0657 & 40.34 & 0.9804 & 0.0691 & 41.13 & 0.9819 & 0.0796 & 9.6 & 17.52\\
EMVD-S \cite{EMVD_CVPR_2020} & 39.71 & 0.9707 & 0.0732 & 39.23 & 0.9735 & 0.0804 & 39.88 & 0.9756 & 0.0875 & 0.81 & 1.66\\
LLRVD \cite{LLRVD_TMM_2022} & 41.51 & 0.9799 & 0.0388 & 41.75 & 0.9823 & 0.0330 & 42.13 & 0.9840 & 0.0350 & 6.29 & 46.14\\
FloRNN \cite{FloRNN_ECCV_2022} & 41.39 & 0.9801 & 0.0468 & 40.74 & 0.9823 & 0.0560 & 41.55 & 0.9842 & 0.0558 & 10.49 & 24.57\\
ShiftNet \cite{ShiftNet_CVPR_2023} & \textbf{42.11} & \textbf{0.9836} & \textbf{0.0328} & \textbf{42.28} & \textbf{0.9848} & \textbf{0.0273} & \textbf{42.70} & \textbf{0.9863} & \textbf{0.0280} & 13.38 & 32.87\\
\hline
BNN \cite{BNN_NIPS_2016} & 33.44 & 0.8417 & 0.2242 & 33.16 & 0.8524 & 0.2961 & 34.25 & 0.8508 & 0.2511 & 0.29 & 1.38 \\
Bireal \cite{Bireal_ECCV_2018} & 36.81 & 0.9227 & 0.1320 & 36.71 & 0.9424 & 0.1572 & 37.34 & 0.9413 & 0.1142 & 0.29 & 1.38 \\
IRNet \cite{IRNet_CVPR_2020} & 33.29 & 0.8119 & 0.1848 & 33.52 & 0.8378 & 0.1834 & 34.35 & 0.8543 & 0.1758 & 0.29 & 1.38 \\
ReActNet \cite{ReActNet_ECCV_2020} & 37.20 & 0.9245 & 0.1502 & 37.27 & 0.9514 & 0.1484 & 37.93 & 0.9449 & 0.1088 & 0.31 & 1.55 \\
BTM \cite{BTM_AAAI_2021} & 37.87 & 0.9445 & 0.0908 & 38.14 & 0.9545 & 0.1081 & 38.14 & 0.9581 & 0.1058 & 0.28 & 1.35 \\
BBCU \cite{BBCU_ICLR_2023} & 39.92 & 0.9736 & 0.0745 & 40.11 & 0.9756 & 0.0660 & 40.48 & 0.9780 & 0.0637 & 0.3 & 1.47 \\
\rowcolor{colorTab}
BRVE (Ours) & \textbf{40.05} & \textbf{0.9742} & \textbf{0.0639} & \textbf{40.25} & \textbf{0.9765} & \textbf{0.0557} & \textbf{40.64} & \textbf{0.9786} & \textbf{0.0570} & 0.3 & 1.49\\
\hline
\end{tabular}
\end{center}
 \vspace{-3mm}
\caption{Quantitative results on SMOID dataset. Our BRVE model outperforms other BNN methods at all three gain levels. BRVE also surpasses the performance of full precision EMVD-S with similar computational costs.}
\label{table:SMOID}
 \vspace{-3mm}
\end{table*}

\paragraph{Spatial Shift.} Both camera motion and object movement in the scene can result in the misalignment of features between frames. Nevertheless, directly fusing the first part feature $\bm{f}_{i}$ with shifted feature $\bm{s}_{i}$ by binary convolutions cannot handle large motions. Following previous work \cite{ShiftNet_CVPR_2023}, we adopt the spatial shift operation to enlarge the receptive field of subsequent binary convolution blocks and cope with misalignment in all possible directions.

Specifically, we first define a spatial shift kernel
\begin{equation}
  \begin{split}
  K = \{ (x, y) | x,y \in \{ -8, -4, 0, 4, 8 \}, (x, y) \neq (0, 0) \},
  \end{split}
\label{eq:ShiftKernel}
\end{equation}
where $|K|=24$ and each element $(x, y)$ in the kernel represents the number of pixels to shift along the corresponding axis. We divide each temporally shifted feature $\bm{s}_{i}$ into $|K|$ slices. Each slice spatially shifts in a different direction according to the shift kernel
\begin{equation}
  \begin{split}
  \bm{s}^{'}_{i,j} = {\rm Shift}(\bm{s}_{i,j}, (x_{j}, y_{j}) ), \  j = 1 ... |K|,
  \end{split}
\label{eq:ShiftOpt}
\end{equation}
where $\bm{s}_{i,j}$ is the $j$-th slice of $\bm{s}_{i}$ and empty area of $\bm{s}^{'}_{i,j}$ caused by spatial shift is filled by zero. Then we concatenate all shifted slices to obtain three spatially shifted features $\{ \bm{s}^{'}_{t-1}, \bm{s}^{'}_{t}, \bm{s}^{'}_{t+1} \}$, where $\bm{s}^{'}_{i} = {\rm Concat}(\bm{s}^{'}_{i,1}, ... ,\bm{s}^{'}_{i,|K|})$.
Finally, we concatenate spatial-temporal shift features together with the unshifted feature for each frame
\begin{equation}
  \begin{split}
  \bm{F}^{s}_{i} = {\rm Concat}(\bm{f}_{i}, \bm{s}^{'}_{i}, \bm{s}_{i}), \  i \in \{ t-1, t, t+1 \},
  \end{split}
\label{eq:SpaTempConcat}
\end{equation}
where $\bm{F}^{s}_{i} \in \mathbb{R}^{H \times W \times 3C/2}$ is then fed into a binary fusion block and several binary conv blocks for spatial-temporal feature fusion, as shown in Figure \ref{fig:framework} \textcolor{red}{(c)}.
\section{Experiments}

In this section, we evaluate our BRVE model on two low-light raw video enhancement datasets. We also conduct extensive experiments to analyze our proposed model.

\subsection{Experiment Settings}

\paragraph{Datasets.} 
Two low-light raw video datasets are adopted in our experiment.
SMOID \cite{SMOID_ICCV_2019} has 309 video pairs of 103 scenes. The resolution of each frame is $480 \times 640$. Three low-light videos are captured in each scene using different ADC gain levels (\ie 0, 15, and 30). We use 70 scenes for training, 16 for validation, and 17 for testing. 
LLRVD \cite{LLRVD_TMM_2022} contains 210 video pairs of 70 scenes. The resolution of each frame is $1400 \times 2600$. For each scene, three videos with different low-light ratios are captured by fixing ISO and adjusting exposure time. We use 60 scenes for training, 4 for validation, and 6 for testing. 
These two datasets are captured with different cameras and SMOID has different noise distribution at different gain levels, which can comprehensively evaluate our BRVE model. \vspace{-4mm}

\paragraph{Training Details.} We use a sequence of 10 frames to train the models on both datasets. The batch size is set to 1 and the input raw patch size is set to $256 \times 256$. We adopt Charbonnier loss \cite{Charbloss_ICIP_1994} to train 100K iterations using Adam optimizer \cite{Adam_ICLR_2015} and the cosine annealing scheduler with the learning rate set to $2 \times 10^{-4}$.
We implement BRVE with PyTorch and train it on an NVIDIA RTX 3090 GPU. \vspace{-4mm}

\begin{table}
\renewcommand\arraystretch{1}
\begin{center}
\begin{tabular}{lccc}
\hline
Method & PSNR $\uparrow$ & SSIM $\uparrow$ & ST-RRED $\downarrow$\\
\hline
SMOID \cite{SMOID_ICCV_2019} & 37.07 & 0.9580 & 0.0391\\
RViDeNet \cite{RViDeNet_CVPR_2020} & 37.39 & 0.9620 & 0.0400\\
FastDVD \cite{FastDVD_CVPR_2020} & 36.39 & 0.9450 & 0.0724\\
EMVD-L \cite{EMVD_CVPR_2020} & 37.00 & 0.9576 & 0.0473\\
EMVD-S \cite{EMVD_CVPR_2020} & 36.70 & 0.9527 & 0.0527\\
LLRVD \cite{LLRVD_TMM_2022} & 37.74 & 0.9650 & 0.0347\\
FloRNN \cite{FloRNN_ECCV_2022} & 37.47 & 0.9634 & 0.0377\\
ShiftNet \cite{ShiftNet_CVPR_2023} & \textbf{37.87} & \textbf{0.9661} & \textbf{0.0346}\\
\hline
BNN \cite{BNN_NIPS_2016} & 31.04 & 0.7393 & 0.0817\\
Bireal \cite{Bireal_ECCV_2018} & 35.90 & 0.9356 & 0.0701\\
IRNet \cite{IRNet_CVPR_2020} & 34.10 & 0.8768 & 0.0967\\
ReActNet \cite{ReActNet_ECCV_2020} & 35.78 & 0.9330 & 0.0697\\
BTM \cite{BTM_AAAI_2021} & 36.56 & 0.9499 & 0.0556\\
BBCU \cite{BBCU_ICLR_2023} & 36.95 & 0.9575 & 0.0457\\
\rowcolor{colorTab}
BRVE (ours) & \textbf{37.07} & \textbf{0.9581} & \textbf{0.0455}\\
\hline
\end{tabular}
\end{center}
 \vspace{-3mm}
\caption{Quantitative results on LLRVD dataset.}
\label{table:LLRVD}
 \vspace{-3mm}
\end{table}

\paragraph{Metrics.} We use the average peak signal-to-noise ratio (PSNR) and structural similarity (SSIM) of all enhanced raw frames to assess the restoration quality of each method. In order to take temporal distortion into account, we use the spatio-temporal reduced reference entropic differences (ST-RRED) \cite{STRRED_TCSVT_2013} to evaluate the video quality. Higher PSNR and SSIM mean better restoration quality and lower ST-RRED indicates better video fidelity. \vspace{-4mm}

\begin{figure*}[]
\small
\centering
  \setlength{\tabcolsep}{0.9mm}
  \renewcommand{\arraystretch}{1}
 \begin{minipage}[t]{0.15\linewidth}
  \begin{tabular}{c}
  \includegraphics[width=1.8\linewidth, height=1.8\linewidth]{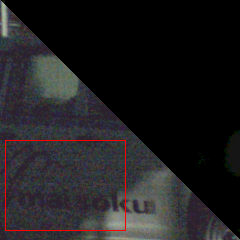}\\
  Low-light / Linear Scaled Frame \\
  \end{tabular}
 \end{minipage}
 \begin{minipage}[t]{0.84\linewidth}
 \flushright
  \begin{tabular}{ccccc}
     \includegraphics[width=0.20\linewidth, height=0.145\linewidth]{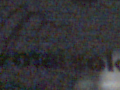}
   & \includegraphics[width=0.20\linewidth, height=0.145\linewidth]{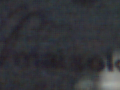}
   & \includegraphics[width=0.20\linewidth, height=0.145\linewidth]{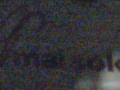}
   & \includegraphics[width=0.20\linewidth, height=0.145\linewidth]{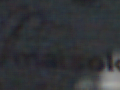}\\
   BNN \cite{BNN_NIPS_2016} & Bireal \cite{Bireal_ECCV_2018} & IRNet \cite{IRNet_CVPR_2020} & ReActNet \cite{ReActNet_ECCV_2020}\\

     \includegraphics[width=0.20\linewidth, height=0.145\linewidth]{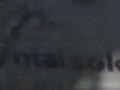}
   & \includegraphics[width=0.20\linewidth, height=0.145\linewidth]{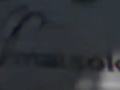}
   & \includegraphics[width=0.20\linewidth, height=0.145\linewidth]{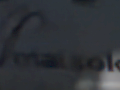}
   & \includegraphics[width=0.20\linewidth, height=0.145\linewidth]{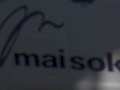}\\
    BTM \cite{BTM_AAAI_2021} & BBCU \cite{BBCU_ICLR_2023} & BRVE (ours) & GT \\
  \end{tabular}
 \end{minipage}

 \vspace{-2mm}
 \caption{Visual comparison of different low-light video enhancement methods on SMOID datasets.}
 \label{fig:SMOID}
 \vspace{-2mm}
\end{figure*}

\begin{figure*}[]
\small
\centering
  \setlength{\tabcolsep}{0.9mm}
  \renewcommand{\arraystretch}{1}
 \begin{minipage}[t]{0.15\linewidth}
  \begin{tabular}{c}
  \includegraphics[width=1.8\linewidth, height=1.8\linewidth]{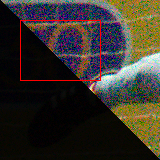}\\
  Low-light / Linear Scaled Frame \\
  \end{tabular}
 \end{minipage}
 \begin{minipage}[t]{0.84\linewidth}
 \flushright
  \begin{tabular}{ccccc}
     \includegraphics[width=0.20\linewidth, height=0.145\linewidth]{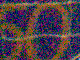}
   & \includegraphics[width=0.20\linewidth, height=0.145\linewidth]{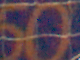}
   & \includegraphics[width=0.20\linewidth, height=0.145\linewidth]{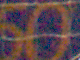}
   & \includegraphics[width=0.20\linewidth, height=0.145\linewidth]{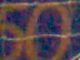}\\
   BNN \cite{BNN_NIPS_2016} & Bireal \cite{Bireal_ECCV_2018} & IRNet \cite{IRNet_CVPR_2020} & ReActNet \cite{ReActNet_ECCV_2020}\\

     \includegraphics[width=0.20\linewidth, height=0.145\linewidth]{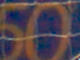}
   & \includegraphics[width=0.20\linewidth, height=0.145\linewidth]{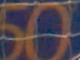}
   & \includegraphics[width=0.20\linewidth, height=0.145\linewidth]{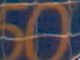}
   & \includegraphics[width=0.20\linewidth, height=0.145\linewidth]{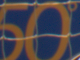}\\
    BTM \cite{BTM_AAAI_2021} & BBCU \cite{BBCU_ICLR_2023} & BRVE (ours) & GT \\
  \end{tabular}
 \end{minipage}

 \vspace{-2mm}
 \caption{Visual comparison of different low-light video enhancement methods on LLRVD datasets.}
 \label{fig:LLRVD}
 \vspace{-2mm}
\end{figure*}

\paragraph{Effciency Evaluation.} Following previous BNN work \cite{XNORNet_ECCV_2016,Bireal_ECCV_2018,BBCU_ICLR_2023,BSCI_arxiv_2023}, we calculate BNN operations by ${\rm OPs}^{b} = {\rm OPs}^{f} / 64$ where ${\rm OPs}^{f}$ is full precision operations, and the parameters of BNNs is computed by ${\rm Params}^{b} = {\rm Params}^{f} / 32$ where ${\rm Params}^{f}$ is real-valued parameters. The overall FLOPs is ${\rm OPs}^{b} + {\rm OPs}^{f}$, we use each model to process a video with 100 frames and a resolution of $256 \times 256$ for counting the per-frame FLOPs. The total number of parameters is ${\rm Params}^{b} + {\rm Params}^{f}$.

\subsection{Compare with State-of-the-arts}

\paragraph{Comparison Methods.} We compare our method with various full precision video denoising networks including SMOID \cite{SMOID_ICCV_2019}, RViDeNet \cite{RViDeNet_CVPR_2020}, FastDVD \cite{FastDVD_CVPR_2020}, EMVD \cite{EMVD_CVPR_2020}, LLRVD \cite{LLRVD_TMM_2022}, FloRNN \cite{FloRNN_ECCV_2022}, and ShiftNet \cite{ShiftNet_CVPR_2023}. We train two versions of the lightweight EMVD model, denoted as EMVD-L (large) and EMVD-S (small). We also compare our distribution-aware binary convolution with other binarization methods, including BNN \cite{BNN_NIPS_2016}, Bireal \cite{Bireal_ECCV_2018}, IRNet \cite{IRNet_CVPR_2020}, ReActNet \cite{ReActNet_ECCV_2020}, BTM \cite{BTM_AAAI_2021}, and BBCU \cite{BBCU_ICLR_2023}. \vspace{-4.6mm}

\paragraph{Quantitative Comparison.}
We illustrate the performance on the SMOID dataset and the model complexity in Table \ref{table:SMOID}. The upper part shows the results of full precision models, while the lower part shows the results of the BNNs. It can be observed that ShiftNet \cite{ShiftNet_CVPR_2023} outperforms other full precision methods, which shows the effectiveness of the shift operation. Besides, we can find that all the BNN methods greatly reduce the parameters and operations. Compared with the state-of-the-art BNN module BBCU \cite{BBCU_ICLR_2023}, our distribution-aware binary convolution improves the performance of BNN at all gain levels with only $1.4\%$ additional computation cost. 

Table \ref{table:LLRVD} demonstrates the quantitative results of the LLRVD dataset. Our method not only surpasses the lightweight full precision EMVD-S model with similar FLOPs by 0.49 dB but also achieves comparable results to the EMVD-L model while utilizing only 9.4\% of its FLOPs and 3.3\% of the parameters.

\vspace{-3mm}

\begin{table}
\begin{center}
\vspace{1.5mm}
\renewcommand\arraystretch{1}
\resizebox{\columnwidth}{!}{
\setlength{\tabcolsep}{2.0mm} 
\begin{tabular}{cc|ccc}
\hline
${\rm mean}(| \cdot |)$     & \makecell[c]{${\rm mean}(\cdot)$ \\ ${\rm std}(\cdot)$}  & PSNR $\uparrow$ & SSIM $\uparrow$ & ST-RRED $\downarrow$ \\
\hline
 \XSolidBrush & \XSolidBrush & 36.95 & 0.9575 & 0.0457 \\
 \XSolidBrush & \Checkmark & 37.02 & 0.9578 & 0.0457 \\
 \Checkmark & \XSolidBrush & 37.01 & 0.9577 & \textbf{0.0450} \\
 
 \hline
 \multicolumn{2}{c|}{w/o shift} & 36.70 & 0.9544 & 0.0517 \\
 \rowcolor{colorTab}
 \multicolumn{2}{c|}{BRVE (ours)} & \textbf{37.07} & \textbf{0.9581} & 0.0455 \\

 \hline
\end{tabular}
}
\end{center}
 \vspace{-5mm}
\caption{Ablation study on the DACA module and shift opeartion.}
\label{table:Ablation}
 \vspace{-4mm}
\end{table}

\paragraph{Visual Comparison.} The qualitative results on the SMOID dataset and LLRVD dataset are illustrated in Figures \ref{fig:SMOID} and \ref{fig:LLRVD}, respectively. We show both low-light and noisy linear scaled frame in the left. Benefiting from the distribution-aware binary convolution, our BRVE exhibits superior visual quality, effectively removing the noise and restoring more details compared to other BNN methods.

\subsection{Analysis of the Proposed Method}

\paragraph{Ablation Study.} We first conduct the ablation study to verify the effect of different channel-wise real-valued statistics on our DACA module. The result is shown in Table \ref{table:Ablation}, where the first column represents whether to use the absolute value, the second column represents whether to use the mean and standard deviation.
We can observe that the performance can be improved after injecting each of these statistics into binary convolution. BRVE model obtains a better result by using all of these statistics to describe the distribution characteristic of full precision activations.

We also perform an ablation study on the spatial-temporal shift operation. As shown in the fourth row of Table \ref{table:Ablation}, the performance declines significantly without the shift operation, which indicates the effectiveness of shift operation in exploiting spatial-temporal self-similarity for low-light raw video denoising. \vspace{-3mm}

\begin{table}
\begin{center}
\renewcommand\arraystretch{1}
\resizebox{\columnwidth}{!}{
\setlength{\tabcolsep}{1.4mm} 
\begin{tabular}{lccc}
\hline
Method & PSNR $\uparrow$ & SSIM $\uparrow$ & ST-RRED $\downarrow$\\
\hline
Raw2Raw+ISP & \textbf{30.46} & \textbf{0.8399} & \textbf{0.2021} \\
Raw2RGB & 27.42 & 0.8131 & 0.2952 \\
RGB2RGB & 24.84 & 0.7988 & 0.5951 \\
\hline
Raw2Raw+ISP+H264 & \textbf{30.19} & \textbf{0.8416} & \textbf{0.1828} \\
RGB2RGB+H264 & 24.75 & 0.8158 & 0.5622 \\
H264+RGB2RGB & 20.03 & 0.6130 & 1.4228 \\
\hline
\end{tabular}
}
\end{center}
\vspace{-4mm}
\caption{Comparison of different low-light video enhancement settings on LLRVD dataset. Metrics are computed in RGB domain.}
\label{table:Raw2RGB}
\vspace{-4mm}
\end{table}

\paragraph{Comparison on Video Enhancement Settings.}
Table \ref{table:Raw2RGB} compares different low-light video enhancement settings, including raw-to-raw (Raw2Raw), raw-to-RGB (Raw2RGB), and RGB-to-RGB (RGB2RGB) on the LLRVD dataset. The raw output of the Raw2Raw setting is converted to RGB domain with ISP. All metrics are evaluated in RGB domain with the ground truth (GT). 

The Raw2Raw setting adopted in this paper shows the best performance, where the low-light enhancement model serves as a pre-processing step before ISP. In the Raw2RGB setting, the network jointly learns to enhance low-light videos and the non-linear process of ISP. It is less effective than the Raw2Raw setting because it is difficult for the network to learn the ISP accurately. In the RGB2RGB setting, we use the network to enhance dark RGB videos, which leads to a significant performance decline. The reason is that RGB videos are quantized into 8-bit and the information of low-intensity pixels in the dark is lost. 

In addition, we also discuss the influence of video compression on low-light video enhancement. As shown in Table \ref{table:Raw2RGB}, H264 compression has little impact on the quality of enhanced videos. However, enhancing compressed low-light videos causes severe degradation, which demonstrates the importance of enhancing the video in the early stages of the image processing pipeline on edge devices. \vspace{-2mm}

\begin{figure}
  \begin{center}
  \includegraphics[width=\linewidth]{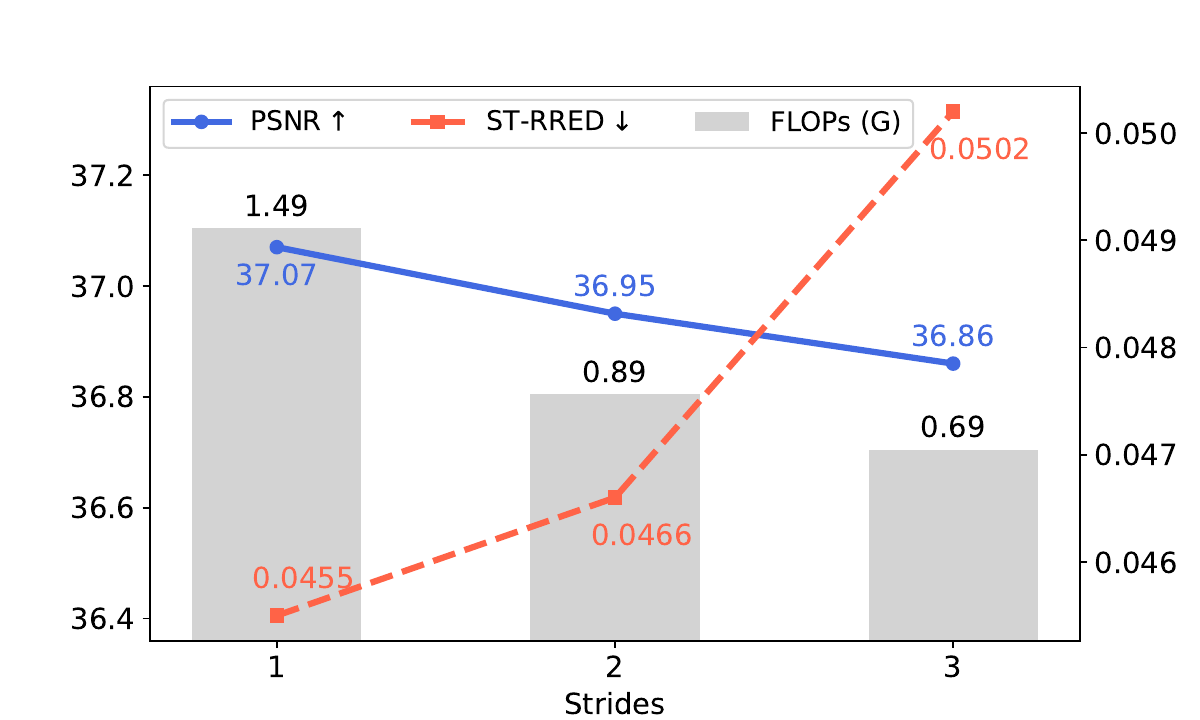}
  \end{center}
  \vspace{-6mm}
\caption{Effect of the sliding window stride on recurrent propagation. A larger sliding stride can further boost the efficiency while compromising the temporal consistency.}
\label{fig:stride}
  \vspace{-5mm}
\end{figure}

\paragraph{Effect of Sliding Stride.} We discuss the influence of the sliding window stride on the recurrent propagation. In the BRVE model, the stride is set to one and the shift binary U-Net outputs one feature at a time. The last two features are used as recurrent embeddings.
We can also set the stride to two or three to further improve the model efficiency. When setting the stride to two, the shift binary U-Net outputs two features at a time and uses the last feature as the recurrent embedding. When setting the stride to three, the shift binary U-Net outputs all three features in the local window, and no recurrent embedding is added between windows.

As shown in Figure \ref{fig:stride}, using a stride of two reduces about 40\% FLOPs compared to the original BRVE model. But the performance drops 0.12 dB in PSNR, which indicates that using more features in recurrent embedding is helpful for video enhancement as more long-term history information is incorporated. Using a stride of three to process each window independently can maximize efficiency. But the large increment of ST-RRED shows that the temporal consistency is greatly impaired without recurrent propagation.

\vspace{-1mm}
\section{Conclusion}
\vspace{-1mm}
In this paper, we propose a BRVE model for low-light raw video enhancement. 
First, we propose the distribution-aware convolution to mitigate the performance gap between BNNs and CNNs. It extracts full precision distribution characteristic information with DACA module to improve the capability of binary convolutions.
Second, we introduce a spatial-temporal shift operation to fuse temporal information and maintain temporal consistency.
It is parameter-free and enables efficient parallel processing for features in a local window.
Experiments on two low-light raw video datasets demonstrate our BRVE model can achieve comparable performance with some full precision models using less memory and computational costs. 

\vspace{-3mm}
\paragraph{Acknowledgement} This work was supported by the National Natural Science Foundation of China (62331006,62171038, 62088101, and 62271414), the R\&D Program of Beijing Municipal Education Commission (KZ202211417048), Science Fund for Distinguished Young Scholars of Zhejiang Province (LR23F010001), and the Fundamental Research Funds for the Central Universities.

{
    \small
    \bibliographystyle{ieeenat_fullname}
    \bibliography{main}
}


\end{document}